\documentclass{article}
\usepackage{microtype}
\usepackage{graphicx}
\usepackage{subfigure}
\usepackage{booktabs} 

\usepackage{hyperref}

\usepackage[accepted]{icml2023}

\usepackage{amsmath}
\usepackage{amssymb}
\usepackage{mathtools}
\usepackage{amsthm}
\usepackage{natbib} 
\usepackage{mathtools} 
\usepackage{bbm} 
\usepackage{amsmath} 
\usepackage{chemformula}   
\usepackage{wrapfig,lipsum,booktabs}   
\usepackage{enumitem}  
\usepackage{amsfonts}   
\usepackage{multirow}   

\usepackage[capitalize,noabbrev]{cleveref}

\theoremstyle{plain}
\newtheorem{theorem}{Theorem}[section]

\theoremstyle{definition}
\newtheorem{definition}[theorem]{Definition}

\usepackage[textsize=tiny]{todonotes}

\icmltitlerunning{Rethinking Explaining Graph Neural Networks via Non-parametric Subgraph Matching}

\begin{document}
\twocolumn[
\icmltitle{Rethinking Explaining Graph Neural Networks via Non-parametric Subgraph Matching}

\begin{icmlauthorlist}
\icmlauthor{Fang Wu}{west,comp}
\icmlauthor{Siyuan Li}{west}
\icmlauthor{Xurui Jin}{comp}
\icmlauthor{Yinghui Jiang}{comp}
\icmlauthor{Dragomir Radev}{yale}
\icmlauthor{Zhangming Niu}{comp}
\icmlauthor{Stan Z. Li}{west}
\end{icmlauthorlist}

\icmlaffiliation{west}{School of Engineering, Westlake University, Hangzhou, China}
\icmlaffiliation{comp}{MindrankAI, Hangzhou, China}
\icmlaffiliation{yale}{Department of Computer Science, Yale University, New Haven, United States}

\icmlcorrespondingauthor{Zhangming Niu}{zhangming@mindrank.ai}
\icmlcorrespondingauthor{Stan Z. Li}{stan.zq.li@westlake.edu.cn}

\icmlkeywords{Machine Learning, ICML}
\vskip 0.3in
]
\printAffiliationsAndNotice{}  

\begin{abstract}
The success of graph neural networks (GNNs) provokes the question about explainability: \emph{``Which fraction of the input graph is the most determinant of the prediction?''} Particularly, parametric explainers prevail in existing approaches because of their more robust capability to decipher the black-box (\emph{i.e.}, target GNNs). 
In this paper, based on the observation that graphs typically share some common motif patterns, we propose a novel non-parametric subgraph matching framework, dubbed MatchExplainer, to explore explanatory subgraphs. It couples the target graph with other counterpart instances and identifies the most crucial joint substructure by minimizing the node corresponding-based distance. 
Moreover, we note that present graph sampling or node-dropping methods usually suffer from the false positive sampling problem. To alleviate this issue, we design a new augmentation paradigm named MatchDrop. It takes advantage of MatchExplainer to fix the most informative portion of the graph and merely operates graph augmentations on the rest less informative part. 
Extensive experiments on synthetic and real-world datasets show the effectiveness of our MatchExplainer by outperforming all state-of-the-art parametric baselines with significant margins. Results also demonstrate that MatchDrop is a general scheme to be equipped with GNNs for enhanced performance. The code is available at~\url{https://github.com/smiles724/MatchExplainer}. 
\end{abstract}
\section{Introduction}
\emph{Graph neural networks} (GNNs) have drawn broad interest due to their success in learning representations of graph-structured data, such as social networks~\citep{fan2019graph}, knowledge graphs~\citep{schlichtkrull2018modeling}, traffic networks~\citep{geng2019spatiotemporal}, and molecular graphs~\citep{gilmer2017neural,wu2023molformer}. Despite their remarkable efficacy, GNNs lack transparency as the rationale of their predictions is not easy for humans to comprehend. This prohibits practitioners from not only gaining an understanding of the network characteristics but correcting systematic patterns of mistakes made by models before deploying them in real-world applications. 

Extensive studies have noticed this issue and great efforts are devoted to explaining GNNs~\citep{yuan2020explainability}. Researchers strive to answer questions like \emph{``What knowledge of the input graph is the most dominantly important in the model's decision?"} To this end, feature attribution and selection~\citep{selvaraju2017grad,sundararajan2017axiomatic,ancona2017towards} becomes a prevalent paradigm. They distribute the model's outcome prediction to the input graph via gradient-like signals~\citep{baldassarre2019explainability,pope2019explainability,schnake2020higher}, mask or attention scores~\citep{ying2019gnnexplainer,luo2020parameterized}, or prediction changes on perturbed features~\citep{schwab2019cxplain,yuan2021explainability}, and then choose a salient substructure as the explanation. 

Apart from them, more recent approaches prefer relying on a deep learning network to parameterize the generation process of explanations~\citep{vu2020pgm,wang2021towards}. These learning-based mechanisms empirically show superior accuracy than the above-mentioned non-parametric ones. Some explainer models are optimized toward local fidelity~\citep{chen2018learning}, such as GNNExplainer~\citep{ying2019gnnexplainer}, PGM-Explainer~\citep{vu2020pgm} and SubgraphX~\citep{yuan2021explainability}. Meanwhile, several others are committed to providing a global understanding of the model prediction, including PGExplainer~\citep{luo2020parameterized}, XGNN~\citep{yuan2020xgnn}, and ReFine~\citep{wang2021towards}. 

Despite the fruitful progress and the popular trend towards parametric explainers, we observe that different essential subgraph patterns are shared by different groups of graphs, which can be the key to deciphering the decision of GNNs. These frequently occurring motifs contain rich semantic meanings and indicate the characteristics of the whole graph instance~\citep{henderson2012rolx,zhang2020motif,banjade2021structure,wu2023molformer}. For example, the hydroxide group (-OH) in small molecules typically results in higher water solubility and a carboxyl group (-COOH) usually contributes to better stability and higher boiling points. Besides that, the pivotal role of functional groups has also been proven in protein structure prediction~\citep{senior2020improved}. 

\begin{figure}
\centering
\includegraphics[scale=0.82]{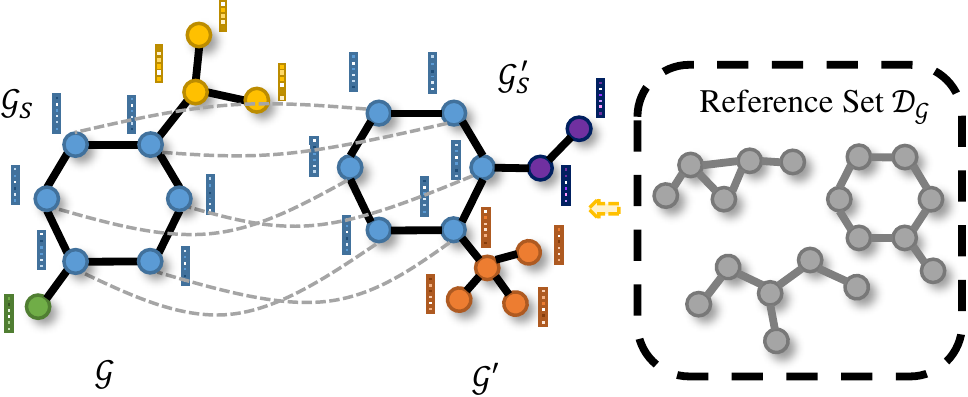}
\vspace{-0.5em}
\caption{The illustration of our proposed MatchExplainer. The explanation $\mathcal{G}_S$ is attained via subgraph matching between $\mathcal{G}$ and $\mathcal{G}'$, where we minimize the accumulated node-to-node distance in the high-dimensional feature space in a greedy search manner. Since several $\mathcal{G}_S$ can be obtained by matching $\mathcal{G}$ to different counterpart graphs $\mathcal{G}'$ from the reference set $\mathcal{D}_{\mathcal{G}}$, we seek to find the optimal one by maximizing Equ.~\ref{equ: subgraph_optimize}. }
\label{fig: matchexplainer}
\vspace{-1em}
\end{figure}

Inspired by this inspection, we propose to mine the explanatory motif in a subgraph matching manner and design a novel non-parametric algorithm dubbed MatchExplainer, whose workflow is depicted in Fig.~\ref{fig: matchexplainer}. 
For each pair of graphs, our MatchExplainer endeavors to explore the most crucial joint substructure by minimizing their node corresponding-based distance in the high-dimensional feature space.
Then it marries the target graph iteratively with other counterpart graphs in the reference set to seek potential explanatory subgraphs. Consequently, unlike traditional explainers, the candidate explanation produced by MatchExplainer can be non-unique for the same target graph instance. 

Taking a step further, we leverage the metric of mutual information from the information theory to analyze the working principle of our MatchExplainer. To be specific, we define the explanation that contains all shared information between paired graphs as \emph{sufficient explanation}, while the explanation that contains the shared and eliminates the non-shared information as \emph{minimal sufficient explanation}. We prove that the minimal sufficient explanation can be used to approximate the desired ground truth explanation with a theoretical guarantee. 
This strong relationship also provides a perspective for us to filter out the best-case substructure from all candidate explanatory subgraphs. To be precise, we propose to optimize the final candidate explanations by maximizing the difference in the prediction after the explanatory subgraph is removed from the original graph.

Last but not least, we exhibit a bonus of our MatchExplainer
to be applied in enhancing the traditional graph augmentation methods. Though exhibiting strong power in preventing over-fitting and over-smoothing, present graph sampling or node-dropping mechanisms suffer from the false positive sampling problem. That is, nodes or edges of the most informative substructure are accidentally dropped or erased but the model is still required to forecast the original property, which can be misleading. To alleviate this obstacle, we take advantage of MatchExplainer and introduce a simple technique called MatchDrop. Specifically, it first digs out the explanatory subgraph by means of MatchExplainer and keeps this part unchanged. Then the graph sampling or node dropping is implemented solely on the remaining less informative part. As a consequence, the core fraction of the input graph that reveals the label information is not affected and the false positive sampling issue is effectively mitigated.      

To summarize, we are the foremost to investigate the explainability of GNNs from the perspective of non-parametric subgraph matching to the best of our knowledge. Extensive experiments on synthetic and real-world applications demonstrate that our MatchExplainer can find the explanatory subgraphs fast and accurately with state-of-the-art performance. Additionally, we empirically show that our MatchDrop, a pragmatic application of MatchExplainer, can serve as an efficient way to promote conventional graph augmentation methods.  

\section{Preliminary and Task Description}
In this section, we begin with the description of the GNN explanation task and briefly review the relevant background of graph matching and graph similarity learning (GSL). 

\paragraph{Explanations for GNNs.} Let $h_Y:\mathcal{G}\rightarrow \mathcal{Y}$ denote the well-trained GNN to be explained, which gives the prediction $\hat{Y}$ to approximate the ground truth $Y$. Without loss of generality, we consider the problem of explaining a graph classification task. Our goal is to find an explainer $h_S: \mathcal{G} \rightarrow \mathcal{G_S}$  that discovers the subgraph $\mathcal{G}_S$ from input graph $\mathcal{G}$ as:
\begin{equation}
\label{problem}
    \min_{h_S}\mathcal{R}(h_Y\circ h_S(\mathcal{G}), \hat{Y}), \textrm{s.t.}  |h_S(\mathcal{G})|\leq K,  
\end{equation}
where $\mathcal{R}(.)$ is the risk function such as a cross-entropy loss or a mean squared error (MSE) loss, and $K$ is a constraint on the size of $\mathcal{G}_S$ to attain a compact explanation. That is, $\mathcal{G}_S$ has at most $K$ nodes.

\paragraph{Graph matching.} As a classic combinatorial problem, graph matching is known in general NP-hard~\citep{loiola2007survey}. They require expensive, complex, and impractical solvers, leading to inexact solutions~\citep{wang2020combinatorial}. Given two different graphs $\mathcal{G}_1 = (\mathcal{V}_1, \mathcal{E}_1)$ and $\mathcal{G}_2 = (\mathcal{V}_2, \mathcal{E}_2)$ with $N_1$ and $N_2$ nodes respectively, the matching between them can be generally expressed by the quadratic assignment programming (QAP) form as~\citep{wang2019learning}:
\begin{equation}
    \min_{\mathbf{T}\in \{0,1\}^{N_1\times N_2}} \textrm{vec}(\mathbf{T})^T \mathbf{K} \textrm{vec}(\mathbf{T}),\, s.t., \mathbf{T}\mathbf{1}= \mathbf{1}, \, \mathbf{T}^T\mathbf{1}= \mathbf{1},
\end{equation}
where $\mathbf{T}$ is a binary permutation matrix encoding the node correspondence, and $\mathbf{1}$ denotes a column vector with all elements to be one. $\mathbf{K}$ is the so-called affinity matrix~\citep{leordeanu2005spectral}, whose elements encode the node-to-node and edge-to-edge affinity between $\mathcal{G}_1$ and $\mathcal{G}_2$.

\paragraph{Graph similarity learning.} GSL is a general framework for graph representation learning that requires reasoning about the structures and semantics of graphs~\citep{li2019graph}. We need to produce the similarity score $s(\mathcal{G}_1, \mathcal{G}_2)$ between them. This similarity $s(.,.)$ is typically defined by either exact matches for full-graph or sub-graph isomorphism~\citep{berretti2001efficient,shasha2002algorithmics}, or some measure of structural similarity such as the graph edit distance~\citep{willett1998chemical,raymond2002rascal}. In our setting, $s(.,.)$ depends entirely on whether these two graphs belong to the same category or share very close properties. Then for $\mathcal{G}_1$ and $\mathcal{G}_2$ with the same type, GSL seeks to maximize the mutual information between their representations with the joint distribution $p(\mathcal{G}_1, \mathcal{G}_2)$ as:
\begin{equation}
\label{graph_similarity_learning}
    \max_{f_1,f_2} I(f_1(\mathcal{G}_1), f_2(\mathcal{G}_2), T), 
\end{equation}
where $f_1$ and $f_2$ are encoding functions. They can share the same parameter (i.e., $f_1=f_2$) or be combined into one architecture. $T$ is the random variable representing the information required for a specific task, independent of the model selection. 

\section{The MatchExplainer Approach}
The majority of recent approaches lean on parametric networks to interpret GNNs, and some early methods for GNN explanations are based on local explainability and from a single-graph view~\citep{ying2019gnnexplainer,baldassarre2019explainability,pope2019explainability,schwab2019cxplain}. Regardless of this inclination, we argue that a non-parametric graph-graph fashion can also excavate important subgraphs and may lead to better explainability.  In this work, we introduce {MatchExplainer} to explain GNNs via identifying the joint essential substructures by means of subgraph matching (see Algorithm~\ref{alg: matchexplainer}).

\subsection{Theoretical Analysis of MatchExplainer}
\label{theory}
From the perspective of probability theory and information theory, Equ.~\ref{problem} is equivalent to maximizing the mutual information between the input graph $\mathcal{G}$ and the subgraph $\mathcal{G}_S$ in the context of $h_Y$. Namely, the goal of an explainer is to derive a small subgraph $\mathcal{G}_S$ such that: 
\begin{equation}
    \max_{\mathcal{G}_S\subset \mathcal{G}, |\mathcal{G}_S|\leq K} I(\mathcal{G}_S, T_h),
\label{equ: mutual_subgraph}
\end{equation}
where $I(.)$ refers to the Shannon mutual information of two random variables. Unlike $T$ which is model-agnostic, $T_h$ represents the knowledge learned by the GNN predictor $h_Y$ in a concrete downstream task. Notably, instead of merely optimizing the information hidden in $\mathcal{G}_S$, another line of research~\citep{yuan2021explainability} seeks to reduce the mutual information between the remaining subgraph $\mathcal{G} - \mathcal{G}_S$ and the original one $\mathcal{G}$ as:
\begin{equation}
    \min_{\mathcal{G}_S\subset \mathcal{G}, |\mathcal{G}_S|\leq K} I(\mathcal{G} - \mathcal{G}_S, T_h).
\label{equ: mutual_subgraph_2}
\end{equation}

As an approximation of directly optimizing Equ.~\ref{equ: mutual_subgraph}, the core idea of MatchExplainer is to fetch another graph $\mathcal{G}'$ that shares the same predicted property as $\mathcal{G}$ (\emph{i.e.}, $h_Y(\mathcal{G}) = h_Y(\mathcal{G}')$) and then extract the most relevant part between them as the explanations. To be specific, we aim to search for the best counterpart $\mathcal{G}'$ so that the mutual information between the input graph $\mathcal{G}$ and the subgraph $\mathcal{G}_S$ is maximized as: 
\begin{equation}
    \max_{\mathcal{G}'\in \mathcal{D}_{\mathcal{G}}, \mathcal{G}'\neq \mathcal{G}} \left[\max_{\mathcal{G}_S\subset \mathcal{G}, |\mathcal{G}_S|\leq K} I(\mathcal{G}_S, \mathcal{G}', T_h)\right],
\end{equation}
where $\mathcal{D}_{\mathcal{G}}$ denotes the reference set consisting of all available graphs, and $\mathcal{G}_S$ is obtained by subgraph matching between $\mathcal{G}$ and $\mathcal{G}'$. 
Similar to the information bottleneck theory~\citep{tishby2015deep,achille2018emergence} in supervised learning, we can define the sufficient explanation and minimal sufficient explanation of $\mathcal{G}$ with its counterpart $\mathcal{G}'\neq \mathcal{G}$ in the context of subgraph matching. 
\begin{definition}[Sufficient Explanation]
Given $\mathcal{G}'$, the explanation $\mathcal{G}^{suf}_S$ of $\mathcal{G}$ is sufficient if and only if $I(\mathcal{G}^{suf}_S, \mathcal{G}', T_{h}) =I(\mathcal{G}, \mathcal{G}', T_{h})$.
\end{definition}
The sufficient explanation $\mathcal{G}^{suf}_S$ of $\mathcal{G}$ keeps all joint information with $\mathcal{G}'$ related to the learned information $T_{h}$. In other words, $\mathcal{G}^{suf}_S$ contains all the shared information between $\mathcal{G}$ and $\mathcal{G}'$. Symmetrically, the sufficient explanation for $\mathcal{G}'$ satisfies $I({\mathcal{G}'}^{suf}_S, \mathcal{G}', T_{h}) =I(\mathcal{G}, \mathcal{G}', T_{h})$. 

\begin{definition}[Minimal Sufficient Explanation]
Given $\mathcal{G}'$, the sufficient explanation $\mathcal{G}^{min}_S$ of $\mathcal{G}$ is minimal if and only if $I(\mathcal{G}^{min}_S, \mathcal{G}, T_{h}) \leq I(\mathcal{G}^{suf}_S, \mathcal{G}, T_{h})$.
\end{definition}
Among all sufficient explanations, the minimal sufficient explanation $\mathcal{G}^{min}_S$ contains the least information about $\mathcal{G}$ with regards to the learned knowledge $T_{h}$. Normally, it is usually assumed that $\mathcal{G}^{min}_S$ only maintains the shared information between $\mathcal{G}$ and $\mathcal{G}'$, and eliminates other non-shared one, i.e., $I(\mathcal{G}^{min}_S, \mathcal{G}|\mathcal{G}')=0$. 

\begin{theorem}[Task Relevant Information in Explanations]
\label{theorem}
~\citep{wang2022rethinking} Given $\mathcal{G}'$, the minimal sufficient explanation $\mathcal{G}^{min}_S$ contains less task-relevant information learned by $h_Y$ from input $\mathcal{G}$ than any other sufficient explanation $\mathcal{G}^{suf}_S$. Formally, we have:
\begin{equation}
\begin{split}
    I(\mathcal{G}, T_{h}) & =I(\mathcal{G}^{min}_S,T_{h}) + I(\mathcal{G}, T_{h}|\mathcal{G}') \\
    &\geq I(\mathcal{G}^{suf}_S,T_{h})=I(\mathcal{G}^{min}_S,T_{h}) + I(\mathcal{G}^{suf}_S, \mathcal{G}, T_{h}|\mathcal{G}') \\
    & \geq  I(\mathcal{G}^{min}_S,T_{h}).
\end{split}
\end{equation}
\end{theorem}
Theorem~\ref{theorem} indicates that the mutual information between $\mathcal{G}$ and $T_{h}$ can be divided into two fractions. One is $\mathcal{G}^{min}_S$, which is the interaction between $\mathcal{G}$ and $\mathcal{G}'$ associated with the learned knowledge $T_{h}$. The other is determined by the disjoint structure of $\mathcal{G}$ and $\mathcal{G}'$ with respect to the learned information $T_{h}$. Our subgraph matching is committed to maximizing $I(\mathcal{G}^{min}_S,T_{h})$, which is the lower bound of $I(\mathcal{G}, T_{h})$. Notably, $I(\mathcal{G}, T_{h}|\mathcal{G}')$ is not completely independent to $I(\mathcal{G}^{min}_S,T_{h})$, but is instead the offset of $I(\mathcal{G}^{min}_S,T_{h})$ to $I(\mathcal{G}, T_{h})$. Hence, if we increase $I(\mathcal{G}^{min}_S,T_{h})$, $I(\mathcal{G}, T_{h}|\mathcal{G}')$ is minimized simultaneously. Consequently,  $I(\mathcal{G}^{min}_S,T_{h})$ can be used to not only improve the lower bound of $I(\mathcal{G}, T_{h})$ but approximate $I(\mathcal{G}, T_{h})$, which is exactly our final explanatory object.  This provides a firm theoretical foundation for our {MatchExplainer} to mine the most explanatory substructure via the subgraph matching approach. 

\subsection{Non-parametric Subgraph Exploration}
\paragraph{Preamble.} It is remarkable that our excavation of explanations through subgraph matching has some significant differences from either graph matching or GSL. On the one hand, graph matching algorithms ~\citep{zanfir2018deep,sarlin2020superglue,wang2020combinatorial,wang2021neural} typically establish node correspondence from a whole graph $\mathcal{G}_1$ to another whole graph $\mathcal{G}_2$. However, we seek to construct partial node correspondence between the subgraph of $\mathcal{G}_1$ and the subgraph of $\mathcal{G}_2$.
On the other hand, GSL concentrates on the graph representations encoded by $f_1$ and $f_2$, as well as the ground truth information $T$ rather than the information $T_h$ learned by the GNN predictor $h_Y$. 

Besides, most existing graph matching architectures~\citep{zanfir2018deep,li2019graph,wang2020combinatorial,papakis2020gcnnmatch,liu2021stochastic} are deep learning-based. They utilize a network to forecast the relationship between nodes or graphs, which has several flaws. For instance, the network needs tremendous computational resources to be trained. More importantly, its effectiveness is unreliable and may fail in certain circumstances if the network is not delicately designed. To overcome these limitations, we employ a non-parametric subgraph matching paradigm, which is totally training-free and fast to explore the most informatively joint substructure shared by any pair of input instances.   

\paragraph{Subgraph matching framework.} We break the target GNN $h_Y$ into two consecutive parts: $h_Y =\phi_G \circ \phi_X$, where $\phi_G$ is the aggregator to compute the graph-level representation and predict the properties, and $\phi_X$ is the feature function to update both the node and edge features. Given a graph $\mathcal{G}$ with node features $\mathbf{h}_i\in \mathbb{R}^{\psi_v}, \forall i \in \mathcal{V}$ and edge features $\mathbf{e}_{ij}\in \mathbb{R}^{\psi_e}, \forall (i,j) \in \mathcal{E}$, the renewed output is calculated as $\{\mathbf{h}'_i\}_{i\in \mathcal{V}}, \{\mathbf{e}'_{ij}\}_{(i,j) \in \mathcal{E}} = \phi_X\left(\{\mathbf{h}_i\}_{i\in \mathcal{V}}, \{\mathbf{e}_{ij}\}_{(i,j) \in \mathcal{E}}\right)$, which is forwarded into $\phi_G$ afterwards. 

As analyzed before, our primary goal is to find a subgraph $\mathcal{G}_S$ with $K$ nodes to maximize $I(\mathcal{G}_S, \mathcal{G}', T_h)$. Due to the hypothesis that the optimal counterpart $\mathcal{G}'$ ought to share the same explanatory substructure as $\mathcal{G}$. Our target is equivalent to optimize $I(\mathcal{G}_S, \mathcal{G}'_S, T_h)$ with $\mathcal{G}_S\subset \mathcal{G}$ and $\mathcal{G}'_S\subset \mathcal{G}'$. There we utilize the node correspondence-based distance $d_G$ as a substitution for measuring $I(\mathcal{G}_S, \mathcal{G}'_S, T_h)$, the shared learned information between $\mathcal{G}_S$ and $\mathcal{G}'_S$. Then given a pair of $\mathcal{G}$ and $\mathcal{G}'$, $d_G$ is defined and minimized as follows:
\begin{equation}
\label{equ: node_distance}
\begin{split}
    \min_{\mathcal{G}_S\subset \mathcal{G}, \mathcal{G}'_S\subset \mathcal{G}'} d_G(\mathcal{G}_S, \mathcal{G}'_S) & =\\
    \min_{\mathcal{G}_S\subset \mathcal{G}, \mathcal{G}'_S\subset \mathcal{G}'} & \left(\min_{\mathbf{T}\in \Pi(\mathcal{G}_S, \mathcal{G}'_S)} \left<\mathbf{T}, \mathbf{D}^{\phi_X}\right>\right),   
\end{split}
\end{equation}
where $\mathbf{D}^{\phi_X}$ is the matrix of all pairwise distances between node features of $\mathcal{G}_S$ and $\mathcal{G}'_S$. Its element is calculated as $\mathbf{D}^{\phi_X}_{ij}=d_X(\mathbf{h}'_i, \mathbf{h}'_j)$ $\forall i\in \mathcal{V}, j\in \mathcal{V}'$, where $d_X$ is the standard vector space similarity such as the Euclidean distance and the Hamming distance. The inner optimization is conducted over $\Pi(.,.)$, which is the set of all matrices with prescribed margins defined as:
\begin{equation}
    \Pi(\mathcal{G}_S, \mathcal{G}'_S) = \left\{\mathbf{T} \in \{0,1\}^{K\times K}\, |\, \mathbf{T}\mathbf{1}= \mathbf{1}, \, \mathbf{T}^T\mathbf{1}= \mathbf{1}\right\}.
\end{equation}
Due to the NP-hard nature of graph matching~\citep{loiola2007survey}, we adopt the greedy strategy to optimize $d_G(\mathcal{G}_S, \mathcal{G}'_S)$ and attain the subgraph $\mathcal{G}_S$. It is worth noting that the greedy algorithm does not guarantee to reach the globally optimal solution~\citep{bang2004greedy}, but can yield locally optimal solutions in a reasonable amount of time with the complexity of $O(K)$. 

After that, we feed $\mathcal{G}_S$ into $h_Y$ and examine its correctness. If $h_Y(\mathcal{G}_S) = h_Y(\mathcal{G})$, then $\mathcal{G}_S$ is regarded as the candidate explanation. Otherwise, $\mathcal{G}_S$ is abandoned since it cannot recover the information required by $h_Y$ to predict $\mathcal{G}$. 

\begin{algorithm}[tb]
   \caption{Workflow of MatchExplainer}
   \label{alg: matchexplainer}
\begin{algorithmic}
   \STATE {\bfseries Input:} target GNN $h_Y$, graph $\mathcal{G}$, reference set $\mathcal{D}_{\mathcal{G}}$
   \STATE Initialize an empty candidate list $\mathcal{D}_{S}$.
   \FOR{$\mathcal{G}'\in \mathcal{D}_{\mathcal{G}}$}
   \STATE $\mathcal{G}_S \leftarrow \min_{\mathcal{G}_S\subset \mathcal{G}, \mathcal{G}'_S\subset \mathcal{G}'} d_G(\mathcal{G}_S, \mathcal{G}'_S)$ in Equ.~\ref{equ: node_distance}
   \IF{$h_Y(\mathcal{G}_S) = h_Y(\mathcal{G})$}
   \STATE add $\mathcal{G}_S$ to $\mathcal{D}_{S}$
   \ENDIF
   \ENDFOR
   \STATE $\mathcal{G}_S^+\leftarrow \max_{\mathcal{G}'\in \mathcal{D}_{\mathcal{S}}, \mathcal{G}'\neq \mathcal{G}} \Delta_{\mathcal{G}}(\mathcal{G}', h_Y)$ in Equ.~\ref{equ: subgraph_optimize}
   \STATE  {\bfseries Return:} $\mathcal{G}_S^+$
\end{algorithmic}
\end{algorithm}

\paragraph{Non-uniqueness of GNN explanations.} Unlike prior learning-based GNN explanation methods~\citep{vu2020pgm,wang2021towards,wang2022reinforced} that generate a unique subgraph $\mathcal{G}_S$ for $\mathcal{G}$, our selection of $\mathcal{G}_S$ varies according to the choice of the counterpart $\mathcal{G}'\in \mathcal{D}_{\mathcal{G}}$. Therefore, MatchExplainer can provide many-to-one explanations for a single graph $\mathcal{G}$ once a bunch of counterparts is given. This offers a new understanding that the determinants for GNNs' predictions are non-unique, and GNNs can gain correct predictions based on several different explanatory subgraphs of the same size. 

\paragraph{Optimization of GNN explanations.} Since our {MatchExplainer} is able to discover a variety of possible explanatory subgraphs, how to screen out the most informative one becomes a critical issue. As indicated in Theorem~\ref{theorem}, $I(\mathcal{G}^{min}_S,T_{h})$ is the lower bound of $I(\mathcal{G},T_{h})$, and their difference $I(\mathcal{G}, T_{h}|\mathcal{G}')$ entirely depends on the selection of the matching counterpart $\mathcal{G}'$. Ideally, $\mathcal{G}'$ ought to share the exact same explanatory substructure with $\mathcal{G}$, i.e., $\mathcal{G}_S = \mathcal{G}'_S$. Meanwhile, $\mathcal{G}$ conditioned on $\mathcal{G}'$ is independent to the learned knowledge $T_{h}$, i.e., $I(\mathcal{G},T_{h}|\mathcal{G}')=0$.
Therefore, there are two distinct principles for selecting the counterpart graphs. 

The first line is to seek $\mathcal{G}'$ that has as close the explanatory subgraph as possible to $\mathcal{G}$. The second line is to ensure that $\mathcal{G}$ conditioned on $\mathcal{G}'$ maintains little information relevant to the learned information $T_{h}$. Nevertheless, without sufficient domain knowledge regarding which substructure is majorly responsible for the graph property, it would be impossible for us to manually select the counterpart graph $\mathcal{G}'$ that satisfies $\mathcal{G}_S \approx \mathcal{G}'_S$. 


As a remedy, we consider optimizing an opposite objective described in Equ.~\ref{equ: mutual_subgraph_2}. That is, we desire to minimize the intersection between $\mathcal{G} - \mathcal{G}_S$ and $T_{h}$, \emph{i.e.}, $I(\mathcal{G} - \mathcal{G}_S, T_h)$. Towards this goal, we remove the extracted subgraph $\mathcal{G}_S$ from $\mathcal{G}$ and aspire to confuse GNNs' predictions on the remaining part $\mathcal{G} - \mathcal{G}_S$. Mathematically, the optimal $\mathcal{G}'$ maximizes the difference between the prediction of the whole graph and the prediction of the graph that is subtracted by $\mathcal{G}_S$. In other words,  we wish to retrieve the best explanation $\mathcal{G}_S^+$ via:
\begin{equation}
\begin{split}
\label{equ: subgraph_optimize}
    \max_{\mathcal{G}'\in \mathcal{D}_{\mathcal{S}}, \mathcal{G}'\neq \mathcal{G}} &\Delta_{\mathcal{G}}(\mathcal{G}', h_Y)  = \\ 
    &\max_{\mathcal{G}'\in \mathcal{D}_{\mathcal{S}}, \mathcal{G}'\neq \mathcal{G}} \left[h_Y^{c^*}(\mathcal{G}) - h_Y^{c^*}(\mathcal{G} - \mathcal{G}_S)\right],    
\end{split}
\end{equation}
where $c^*$ is the ground truth class of $\mathcal{G}$ and $\mathcal{G}_S$ is the substructure via subgraph matching with $\mathcal{G}'$. $\mathcal{D}_{\mathcal{S}}$ is the candidate subgraph set. 

To summarize, given any graph $\mathcal{G}$ and a reference graph set $\mathcal{D}_{\mathcal{G}}$, we first acquire all possible subgraphs via matching $\mathcal{G}$ to available counterparts in $\mathcal{D}_{\mathcal{G}}$. After the pairwise subgraph matching, we calculate their corresponding $\Delta_{\mathcal{G}}(., h_Y)$ and pick up the one that leads to the largest $\Delta_{\mathcal{G}}(., h_Y)$ as the optimal counterpart graph. Notably, not all graphs in $\mathcal{D}_{\mathcal{G}}$ are qualified counterparts and there are several intuitive conditions that $\mathcal{G}'$ has to satisfy. First, $\mathcal{G}$ and $\mathcal{G}'$ should belong to the same category predicted by $h_Y$. Besides, $\mathcal{G}'$ needs to have at least $K$ nodes. Otherwise, $G_S$ would be smaller than the given constrained size. 

\paragraph{Effectiveness vs. efficiency.} The time-complexity is always a vital topic to evaluate the practicability of explainers. For our MatchExplainer, the size of the reference set, \emph{i.e.}, $|\mathcal{D}_{\mathcal{G}}|$, plays a vital role in determining the time cost since the total time cost is $O(K|\mathcal{D}_{\mathcal{G}}|)$. However, a limited number of counterpart graphs can also prohibit it from exploring better explanatory subgraphs. Thus, it is non-trivial to balance the effectiveness and efficiency of MatchExplainer by choosing an appropriate size of $\mathcal{D}_{\mathcal{G}}$.

\section{The MatchDrop Methodology}
\paragraph{Prevention of the false positive sampling.} Deep graph learning faces unique challenges, such as feature data incompleteness, structural data sparsity, and over-smoothing. To address these issues, a growing number of data augmentation techniques~\citep{hamilton2017inductive,rong2019dropedge} have been proposed in the graph domain and shown promising outcomes. Graph sampling and node dropping~\citep{feng2020graph,xu2021infogcl} are two commonly used mechanisms. 
However, most previous approaches are completely randomized, resulting in false positive sampling and injecting spurious information into the training process. For instance, \emph{1,3-dinitrobenzene} (\ch{C6H4N2O4}) is a mutagen molecule and its explanation is the \ch{NO2} groups~\citep{debnath1991structure}. If any edge or node of the \ch{NO2} group is accidentally dropped or destroyed, the mutagenicity property no longer exists. Therefore, it will misguide GNNs if the original label is assigned to this molecular graph after node or edge sampling. 

\begin{figure}
\centering
\includegraphics[scale=0.43]{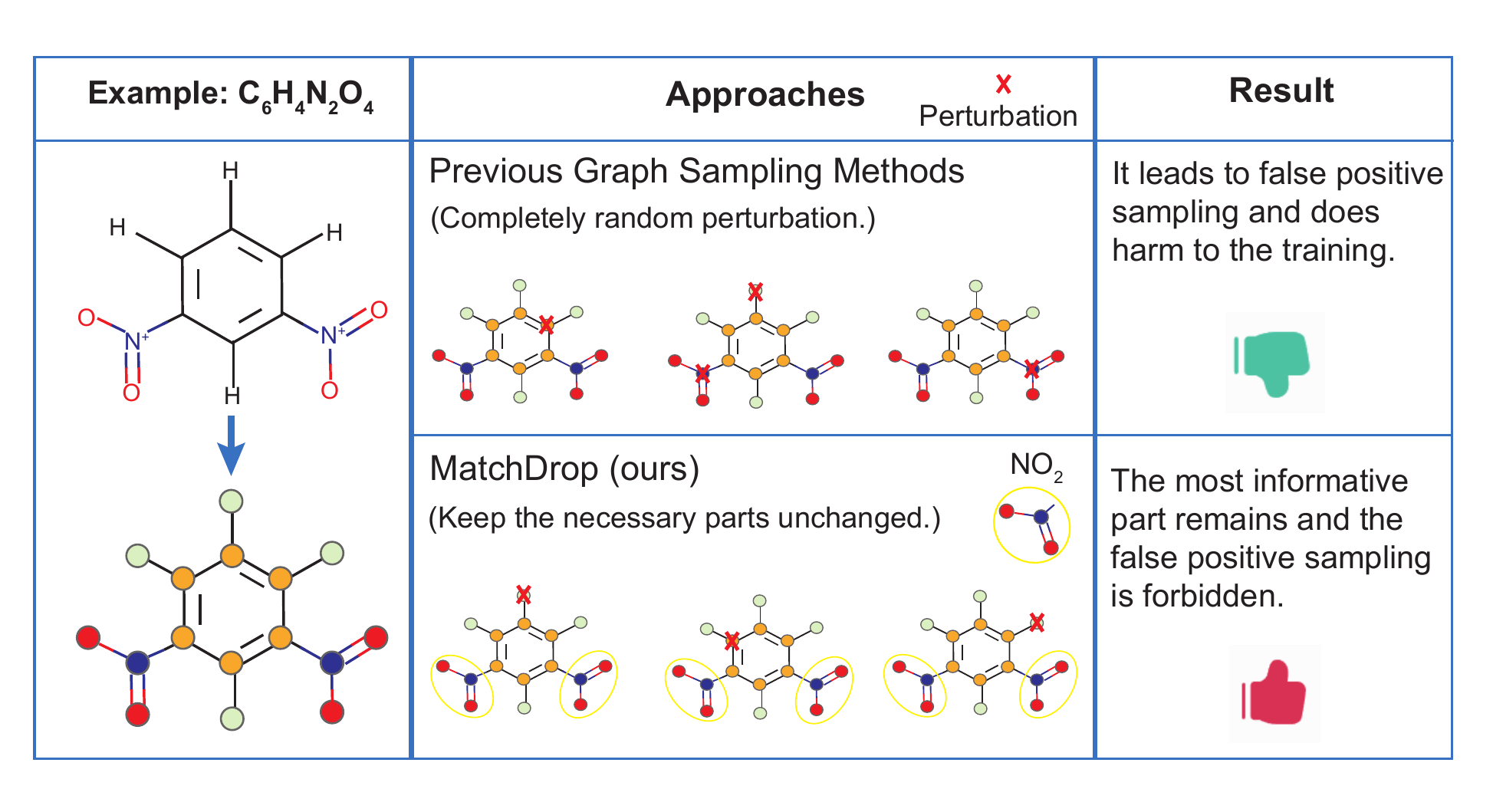}
\vspace{-1.5em}
\caption{Comparison between graph augmentation with and without MatchDrop.}
\label{matchdrop}
\vspace{-1em}
\end{figure}
To tackle this drawback, recall that our {MatchExplainer} offers a convenient way to discover the most essential part of a given graph. It is natural to keep this crucial portion unchanged and only drop nodes or edges in the remaining portion. Based on this idea, we propose a simple but effective method dubbed {MatchDrop}, which keeps the most informative part of graphs found by our {MatchExplainer} and alters the less informative part (see Fig.~\ref{matchdrop}). 

The procedure of our MatchDrop is described as follows. To begin with, we train a GNN $h_Y$ for several epochs until it converges to an acceptable accuracy, which guarantees the effectiveness of the subsequent subgraph selection. Then for each graph $\mathcal{G}$ in training set $\mathcal{D}_{\textrm{train}}$, we randomly select another graph $\mathcal{G}'\in \mathcal{D}_{\textrm{train}}$ with the same class as the counterpart graph. Afterwards, we explore its subgraph $\mathcal{G}_S$ via {MatchExplainer} with a retaining ratio $\rho$  (i.e., $|\mathcal{G}_S| = \rho|\mathcal{G}|$) and use it as the model input to train $h_Y$. 

Notably, similar to the typical image augmentation skills such as rotation and flapping~\citep{shorten2019survey}, MatchDrop is a novel data augmentation technique for GNN training. However, instead of randomly augmenting $\mathcal{G}$, {MatchDrop} reserves the most informative part and only changes the less important substructure. This significantly reduces the possibility of false positive sampling. Additionally, unlike other learnable mechanisms to inspect subgraphs, our MatchDrop is entirely parameter-free and, therefore, can be deployed at any stage of the training period.

\paragraph{Training objective.} The training of GNNs is supervised by the cross entropy (CE) loss. Suppose there are $M$ classes in total, then the loss takes the following form:
\begin{equation}
    \mathcal{L}_{S} = -\frac{1}{|\mathcal{D}_{\textrm{train}}|}\sum_{\mathcal{G} \in \mathcal{D}_{\textrm{train}}} \sum_{c=1}^M {Y}_\mathcal{G} \log \left(h_Y^c\left(h_S(\mathcal{G}, \rho)\right)\right),
\end{equation}
where $h_Y^c(.)$ indicates the predicted probability of $\mathcal{G}_S$ to be of class $c$ and $Y_G$ is the ground truth value. $h_S$ employs {MatchExplainer} to mine the subgraph $\mathcal{G}_S$ by matching $\mathcal{G}$ to a randomly selected counterpart graph $\mathcal{G}'$ in the training set $\mathcal{D}_{\textrm{train}}$ with a pre-defined ratio $\rho$. 

\begin{table*}[t]
\centering
\resizebox{1.65 \columnwidth}{!}{%
\begin{tabular}{lccccc}    
    \toprule & MUTAG & VG-5 & MNIST & \multicolumn{2}{c}{BA-3Motif} \\
    \cline{5-6} & ACC-AUC & ACC-AUC & ACC-AUC & ACC-AUC & Recall@ 5 \\ \midrule 
    SA & $0.769$ & $0.769$ & $0.559$ & $0.518$ & $0.243$ \\
    Grad-CAM & $0.786\pm0.011$ & $0.909\pm0.005$ & $0.581\pm0.009$ &  $0.533\pm0.003$ &  $0.212\pm0.002$\\ 
    GNNExplainer & $0.895\pm0.010$ & $0.895\pm0.003$ & $0.535\pm0.013$ & $0.528\pm0.005$ & $0.157\pm0.002$ \\
    PG-Explainer & $0.631\pm0.008$ & $0.790\pm0.004$ & $0.504\pm0.010$ & $\underline{0.586\pm0.004}$ & $0.293\pm0.001$ \\
    PGM-Explainer & $0.714\pm0.007$ & $0.792\pm0.001$ & $0.615\pm0.003$ & $0.575\pm0.002$ & $0.250\pm0.000$ \\
    ReFine & $\underline{0.955}\pm0.005$ & $\underline{0.914}\pm0.001$ & $\underline{0.636}\pm0.003$ & ${0.576}\pm0.013$\footnotemark[1] & $\underline{0.297}\pm0.000$\footnotemark[1] \\ \midrule 
    MatchExplainer &  \textbf{0.997} & \textbf{0.993} & \textbf{0.938} & \textbf{0.634} & \textbf{0.305}  \\ 
    Relative Impro. & $4.5\%$ & $8.6\%$ & $48.9\%$ & $8.1\%$ & $2.6\%$ \\ \bottomrule
\end{tabular}}
\caption{Comparisons of our {MatchExplainer} with other baseline explainers.}
\label{tab:explain_performance}
\vspace{-1em}
\end{table*}
\footnotetext[1]{These results are reproduced}
\section{Experimental Analysis}
\subsection{Datasets and Experimental Settings}
Following~\citet{wang2021towards}, we use four standard datasets with various target GNNs. 

\begin{itemize}[leftmargin=*]
    \item \textbf{Molecule graph classification}: {MUTAG}~\citep{debnath1991structure,kazius2005derivation} is a molecular dataset for the graph classification problem. Each graph represents a molecule with nodes for atoms and edges for bonds. The labels are determined by their mutagenic effect on a bacterium. The well-trained Graph Isomorphism Network (GIN)~\citep{xu2018powerful} has approximately achieved 82\% testing accuracy. 
    \item \textbf{Motif graph classification.}:~\citet{wang2021towards} create a synthetic dataset, BA-3Motif, with 3000 graphs. They use the Barabasi-Albert (BA) graphs as the base and attach each base with one of three motifs: house, cycle, or grid. We train an ASAP model~\citep{ranjan2020asap} that realizes a 99.75\% testing accuracy. 
    \item \textbf{Handwriting graph classification}:~\citet{knyazev2019understanding} transforms the MNIST images into 70K superpixel graphs with at most 75 nodes for each graph. The nodes are superpixels, and the edges are the spatial distances between them. There are ten types of digital labels. We adopt a Spline-based GNN~\citep{fey2018splinecnn} that gains around 98\% accuracy in the testing set.
    
    \item \textbf{Scene graph classification}:~\citet{wang2021towards} select 4443 pairs of images and scene graphs from Visual Genome~\citep{krishna2017visual} to construct the VG-5 dataset~\citep{pope2019explainability}. Each graph is labeled with one of five categories: stadium, street, farm, surfing, and forest. The regions of objects are represented as nodes, while edges indicate the relationships between object nodes. We train an AAPNP~\citep{klicpera2018predict} that reaches 61.9\% testing accuracy. 
\end{itemize}

We compare our {MatchExplainer} with several state-of-the-art and popular explanation baselines, which are listed below:
\begin{itemize}[leftmargin=*]
    \item \textbf{SA}~\citep{baldassarre2019explainability} directly uses the gradients of the model prediction concerning the adjacency matrix of the input graph as the importance of edges. 
    \item \textbf{Grad-CAM}~\citep{selvaraju2017grad,pope2019explainability} uses the gradients of any target concept, such as the motif in a graph flowing into the final convolutional layer, to produce a coarse localization map highlighting the critical regions in the graph for predicting the concept.
    \item \textbf{GNNExplainer}~\citep{ying2019gnnexplainer} optimizes soft masks for edges and node features to  maximize the mutual information between the original predictions and new predictions. 
    \item \textbf{PGExplainer}~\citep{luo2020parameterized} hires a parameterized model to decide whether an edge is essential, which is trained over multiple explained instances with all edges. 
    \item \textbf{PGM-Explainer}~\citep{vu2020pgm} collects the prediction change on the random node perturbations and then learns a Bayesian network from these perturbation-prediction observations to capture the dependencies among the nodes and the prediction. 
    \item \textbf{ReFine}~\citep{wang2021towards} exploits the pre-training and fine-tuning idea to develop a multi-grained GNN explainer. It has a global understanding of model workings and local insights on specific instances.
\end{itemize}

As the ground-truth explanations are usually unknown, it is tough to evaluate the excellence of explanations quantitatively. There, we follow~\citet{wang2021towards} and employ \textbf{the predictive accuracy (ACC@$\boldsymbol{\rho}$)} and \textbf{Recall@$\boldsymbol{N}$} as the metrics. Specifically, ACC@$\rho$ measures the fidelity of the explanatory subgraphs by forwarding them into the target model and examining how well it recovers the target prediction. ACC-AUC is reported as the area under the ACC curve over different selection ratios $\rho\in \{0.1, 0.2, ..., 1.0\}$. Recall@${N}$ is computed as $\mathbb{E}_{\mathcal{G}}\left[\left|\mathcal{G}_{s} \cap \mathcal{G}_{S}^{*}\right| /\left|\mathcal{G}_{S}^{*}\right|\right]$, where $\mathcal{G}_{S}^{*}$ is the ground-truth explanatory subgraph. Remarkably, Recall@${N}$ is only suitable for BA3-motif since this dataset is synthetic and the motifs are foregone. 

\subsection{Can MatchExplainer Find Better Explanations?}
\paragraph{Quantitative results.} To investigate the effectiveness of MatchExplainer, we conduct broad experiments on four datasets, and the comparisons are reported in Table~\ref{tab:explain_performance}. For MUTAG, VG-5, and BA3-Motif, we exploit the full training and validation data as the reference set. For MNIST, we randomly select 10\% available samples as the reference set to speed up matching. It can be found that MatchExplainer outperforms every baseline in all cases. Particularly, previous explainers fail to explain GNNs well in MNIST with ACC-AUCs lower than 65\%, but MatchExplainer can reach as high as 93.8\%. And if we use the whole training and validation data in MNIST as the reference, its ACC-AUC can increase to 97.2\%. This phenomenon demonstrates the advantage of subgraph matching in explaining GNNs when the dataset has clear patterns of explanatory subgraphs. Additionally, MatchExplainer also achieves significant relative improvements over the strongest baseline by 8.6\% and 8.1\% in VG-5 and BA3-Motif, respectively. 

Furthermore, it is also worth noting that MatchExplainer realizes nearly 100\% ACC-AUCs in each task but BA-3Motif. For BA-3Motif, we find that its predictive accuracy are $[0.31, 0.31, 0.31, 0.34, 0.49, 0.71, 0.97, 1.0, 1.0, 1.0]$ with different selection ratios. This aligns with the fact that most motifs in this task occupy a large fraction of the whole graph. Once the selection ratio is greater than 0.7, MatchExplainer can figure out the correct explanatory subgraph.  

\paragraph{Visualization.} In addition, we envision the explanations of MatchExplainer on MUTAG in Appendix~\ref{app_visual} for qualitative evaluations. We also compare the efficiency of our MatchExplainer with other parametric methods in Appendix~\ref{tab: time}. It can be discovered that MatchExplainer enjoys a competitive fast inference speed with no additional training cost, making it possible for large-scale deployment. 

\begin{table*}[t]
\centering
\resizebox{1.6 \columnwidth}{!}{%
\begin{tabular}{cc|ccccc}   \toprule 
    Dataset & Backbone & Original & FPDrop  & DropNode & PGDrop &  {MatchDrop} \\ \midrule
    \multirow{2}{*}{\begin{tabular}[c]{@{}c@{}}MUTAG  \end{tabular}} & GCN & $0.828\pm0.004$ & $0.803\pm0.017$ & $\underline{0.832\pm0.008}$ & $0.825\pm0.02$ & \textbf{0.844$\pm$0.006} \\
    & GIN & $0.832\pm0.003$ & $0.806\pm0.020$ & $\underline{0.835\pm0.009}$ & $0.828\pm0.01$ & \textbf{0.845$\pm$0.007}\\ \midrule 
    \multirow{2}{*}{\begin{tabular}[c]{@{}c@{}}VG-5 \end{tabular}} & GCN & $0.619\pm0.003$ & $0.587\pm0.014$ & $\underline{0.623\pm0.007}$ & $0.604\pm0.002$ & \textbf{0.638$\pm$0.008}\\ 
    & GIN &  $0.621\pm0.004$ & $0.593\pm0.018$ & $\underline{0.622\pm0.006}$ & $0.600\pm0.004$ & \textbf{0.630$\pm$0.003}  \\ \midrule 
    \multirow{2}{*}{\begin{tabular}[c]{@{}c@{}}MNIST  \end{tabular}}  & GCN & $0.982\pm0.001$ & $0.955\pm0.008$ & $\underline{0.982\pm0.002}$ & $0.975\pm0.003$ & \textbf{0.986$\pm$0.002} \\
    & GIN &  $0.988\pm0.001$ & $0.959\pm0.005$ & $\underline{0.989\pm0.001}$ & $0.979\pm0.002$ & \textbf{0.990$\pm$0.001} \\  \bottomrule
\end{tabular}}
\caption{Testing accuracy (\%) comparisons on different backbones with and without {MatchDrop}.}
\label{tab:drop_performance}
\end{table*}
\subsection{Can MatchDrop Improve the Performance of GNNs?}
\paragraph{Implementations.} We take account of two backbones: GCN~\citep{kipf2016semi}, and GIN~\citep{xu2018powerful} with a depth of 6. Similar to~\citet{rong2019dropedge}, we adopt a random hyper-parameter search for each architecture to enable more robust comparisons. There, \emph{DropNode} stands for randomly sampling subgraphs, which can be also treated as a specific form of node dropping. False-positive drop (\emph{FPDrop}) is the opposite operation of our MatchDrop, where the subgraph sampling or node dropping is only performed in the explanatory subgraphs while the rest remains the same. We add FPDrop as a baseline to help unravel the reason why MatchDrop works. \emph{PGDrop} is similar to MatchDrop, but uses a fixed PGExplainer~\citep{luo2020parameterized} to explore the informative substructure. The selection ratios $\rho$ for FPDrop, PGDrop, and MatchDrop are all set as 0.95.

\paragraph{Overall results.} Table~\ref{tab:drop_performance} documents the performance on all datasets except BA-3Motif, since its testing accuracy has already approached 100\%. It can be observed that {MatchDrop} consistently promotes the testing accuracy for all cases. Exceptionally, FPdrop imposes a negative impact over the performance of GNNs. This indicates that false positive sampling does harm to the conventional graph augmentation methods, which can be surmounted by our MatchDrop effectively. On the other hand, PGDrop also gives rise to the decrease of accuracy. One possible reason is that parameterized explainers like PGExplainr are trained on samples that GNNs predict correctly, so they are incapable to explore explanatory subgraphs on unseen graphs that GNNs forecast mistakenly.  

\section{Related Work}
\subsection{Explainability of GNNS}
Interpretability and feature selection have been attached to growing significance in demystifying complicated deep learning models, and increasing interests have been appealed in explaining GNNs~\citep{ying2019gnnexplainer,Wu2022Bottleneck}. Despite fruitful progress, the study in this area is still insufficient compared to the domain of images and natural languages. Generally, there are two mainstream lines of research. The widely-adopted one nowadays is the parametric explanation method. They run a parameterized model to dig out informative substructures or generate the saliency maps. For example, GNNExplainer~\citep{ying2019gnnexplainer} learns soft masks for each instance and applies them to the adjacency matrix. PGExplainer~\citep{luo2020parameterized} collectively explains multiple samples with a probabilistic graph generative model. XGNN~\citep{yuan2020xgnn} utilizes a graph generator to output class-wise graph patterns to explain GNNs for each class. PGM-Explainer~\citep{vu2020pgm} proposes a Bayesian network on the pairs of graph perturbations and prediction changes. The other line is the non-parametric explanation methods, which do not involve any additional trainable models. They employ some heuristics like gradient-like scores obtained by backpropagation as the feature contributions of a specific instance~\citep{baldassarre2019explainability,pope2019explainability,schnake2020higher}. As mentioned, the latter is usually less favored because their performance is much poorer than the former parametric methods. In contrast, our MatchExplainer procures state-of-the-art results astonishingly. 

\subsection{Graph Augmentations}
Data augmentation has recently attracted growing attention in graph representation learning to counter issues like data noise and data scarcity~\citep{zhao2022graph}. The related work can be roughly broken down into \emph{feature-wise}~\citep{zhang2017mixup,liu2021local,taguchi2021graph}, \emph{structure-wise}~\citep{you2020graph,zhao2021data}, and \emph{label-wise}~\citep{verma2019manifold} categories based on the augmentation modality~\citep{ding2022data}. Among them, many efforts are raised to augment the graph structures. Compared with adding or deleting edges~\citep{xu2022graph}, the augmentation operations on node sets are more complicated. A typical application is to promote the propagation of the whole graph by inserting a supernode~\citep{gilmer2017neural}, while~\citet{zhao2021graphsmote} interpolate nodes to enrich the minority classes.
On the contrary, some implement graph or subgraph sampling by dropping nodes for different purposes, such as scaling up GNNs~\citep{hamilton2017inductive}, enabling contrastive learning~\citep{qiu2020gcc}, and prohibiting over-fitting and over-smoothing~\citep{rong2019dropedge}. Nonetheless, few of those graph sampling or node dropping approaches manage to find augmented graph instances from the input graph that best preserve the original properties. 

\section{Conclusion}
This paper proposes a promising subgraph matching technique called MatchExplainer for GNN explanations. Distinct from the popular trend of using a parameterized network that lacks interpretability, we design a non-parametric algorithm to search for the most informative joint subgraph between a pair of graphs with theoretical guarantees. Furthermore, we combine MatchExplainer with the classic graph augmentation method and show its great capacity in ameliorating the false positive sampling challenge. Experiments convincingly demonstrate the efficacy of our MatchExplainer by winning over parametric approaches with significant margins. Our work hopes to push the frontier of non-parametric methods to explain deep learning models.

\bibliography{cite}
\bibliographystyle{icml2023}

\newpage
\appendix
\onecolumn


\section{The Greedy Algorithm for Subgraph Matching}
Here we provide the pseudo-code of subgraph matching in a greedy way. Given two graphs $\mathcal{G}$ and $\mathcal{G}'$, we first feed them to a GNN $h_Y$ and obtain their corresponding node features as $\{\mathbf{h}_i\}_{i\in \mathcal{V}}$ and $\{\mathbf{h}'_i\}_{i\in \mathcal{V'}}$. Our goal is to find subgraphs $\mathcal{G}_S$ and $\mathcal{G}'_S$ such that $d_G(\mathcal{G}_S, \mathcal{G}'_S)$ is minimized. 
\begin{algorithm}[ht]
   \caption{Greedy Algorithm to Explore Shared Subgraphs}
   \label{alg:greedy}
\begin{algorithmic}
   \STATE {\bfseries Input:} node features $\{\mathbf{h}_i\}_{i\in \mathcal{V}}$ and $\{\mathbf{h}'_i\}_{i\in \mathcal{V'}}$, subgraph size $K$
   \STATE Initialize an empty list $g$ to store the selected nodes.
   \STATE Compute the distance matrix $\mathbf{D}^{\phi_X}$, where $\mathbf{D}^{\phi_X}_{ij}=d_X(\mathbf{h}'_i, \mathbf{h}'_j) \forall i\in \mathcal{V}, \forall j\in \mathcal{V}'$.
   \FOR{$t=1,...,K$}
   \STATE $(i,j) \leftarrow \min_{i\in \mathcal{V}, j\in \mathcal{V}'} \mathbf{D}^{\phi_X}_{ij}$,
   \STATE add $i$ to $g$,
   \STATE remove $i$ from $\mathcal{V}$,
   \STATE remove $j$ from $\mathcal{V}'$,
   \ENDFOR
   \STATE  {\bfseries Return:} $g$
\end{algorithmic}
\end{algorithm}

\section{Experimental Details and Additional Results}
\paragraph{Explaining GNNs.}
All experiments are conducted on a single A100 PCIE GPU (40GB). For the parametric methods containing GNNExplainer, PGExplainer, PGM-Explainer, and Refine, we use the reported performance in~\citet{wang2021towards}. Regarding the re-implementation of Refine in BA-3Motif, we use the original code with the same hyperparameters, and we adopt Adam optimizer~\citep{kingma2014adam} and set the learning rate of pre-training and fine-tuning as 1e-3 and 1e-4,
respectively. 

\paragraph{Graph augmentations.} All experiments are also implemented on a single A100 PCIE GPU (40GB). We employ three sorts of different GNN variants (GCN, GAT, and GIN) to fit these datasets and verify the efficacy of various graph augmentation methods. We employ Adam optimizer for model training. For MUTAG, the batch size is 128, and the learning rate is 1e-3. For BA3-Motif, the batch size is 128, and the learning rate is 1e-3. For VG-5, the batch size is 256, and the learning rate is 0.5 * 1e-3. We fix the number of epochs to 100 for all datasets. 

\paragraph{Efficiency studies.} We compute the average inference time cost for each dataset with different methods to obtain explanations of a single graph. We also count the overall training and inference time expenditure and summarize the results in Table~\ref{tab: time}. Specifically, we train GNNExplainer and PG-Explainer for 10 epochs, and pre-train ReFine for 50 epochs before evaluation. It can be observed that though prior approaches enjoy fast inference speed, they suffer from long-term training phases. As an alternative, our MatchExplainer is completely training-free. When comparing the total time, MatchExplainer is the least computationally expensive in MUTAG, VG-5, and MNIST. However, as most motifs in BA-3Motif are large-size, MatchExplainer has to traverse a large reference set to obtain appropriate counterpart graphs, which unavoidably results in spending far more time. 
\begin{table}[h]
\caption{Efficiency studies of different methods (in seconds).}
\label{tab: time}
\centering
\resizebox{0.7 \columnwidth}{!}{%
\begin{tabular}{cc|cccc}   \toprule 
    Method & Phase & MUTAG & VG-5  & MNIST & BA-3Motif \\ \midrule
    \multirow{3}{*}{\begin{tabular}[c]{@{}c@{}}GNNExplainer \end{tabular}} & Training & 186.0 & 1127.2 & 1135.4 & 66.1  \\
    & Inference (per graph) & 1.290 & 0.565 & 0.732 & 0.517 \\
    & Training + Inference (total) & 703.4 & 1644.6 & 1782.1  & {271.6} \\ \midrule 
    \multirow{3}{*}{\begin{tabular}[c]{@{}c@{}}PG-Explainer \end{tabular}} & Training & 186.3 & 286.3 & 1154.1 & 112.4 \\
    & Inference (per graph) & 0.056 & 0.094 & 0.025 & 0.020  \\
    & Training + Inference (total) & \underline{208.6} & \underline{309.5} & \underline{1162.1} & \textbf{120.4} \\ \midrule
    \multirow{3}{*}{\begin{tabular}[c]{@{}c@{}}ReFine \end{tabular}} & Training & 1191.6 & 1933.3 & 5025.8  & 763.0 \\
    & Inference (per graph) & 0.068 & 0.107 & 0.026 & 0.027 \\
    & Training + Inference (total) & 1218.9 & 1959.7 & 5051.2 & 773.8\\ \midrule 
    \multirow{3}{*}{\begin{tabular}[c]{@{}c@{}}MatchExplainer \end{tabular}} & Training & -- & -- & -- & --  \\
    & Inference (per graph) & 0.485 & 0.732 & 0.682 & 7.687 \\
    & Training + Inference (total) & \textbf{194.6} & \textbf{180.3} & \textbf{667.8} & \underline{224.7} \\  \bottomrule
\end{tabular}}
\end{table}

\section{Explanations for Graph Classification Models}
\label{app_visual}
\begin{figure*}[h]
\centering
\includegraphics[scale=0.35]{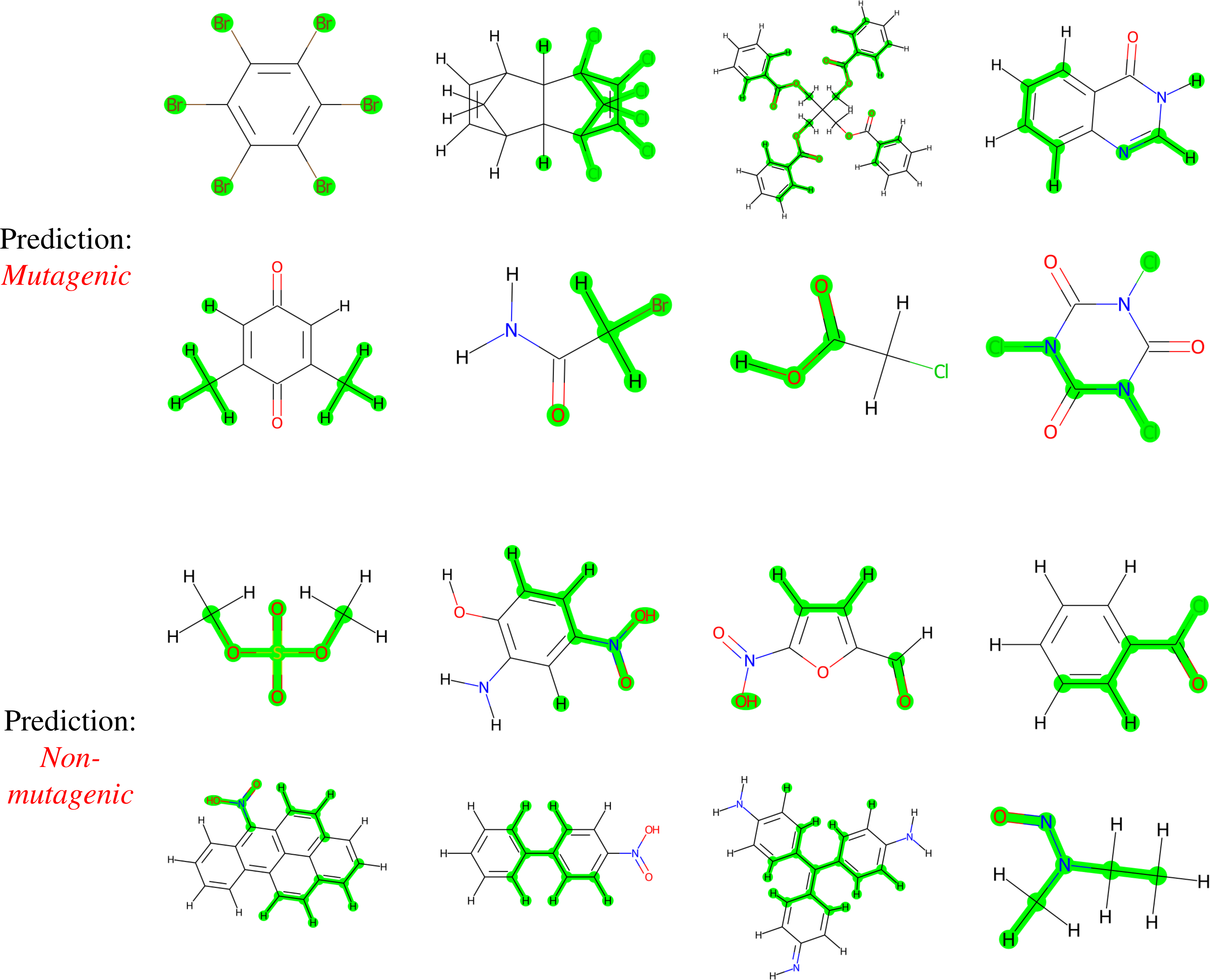}
\caption{Explanatory subgraphs in Mutagenicity, where 50\% nodes are highlighted. }
\label{mutag_visual}
\end{figure*}

In this section, we report visualizations of explanations in Figure~\ref{mutag_visual}. 

\end{document}